\documentclass[11pt,a4paper]{article}
\usepackage[hyperref]{emnlp-ijcnlp-2019}
\usepackage{times}
\usepackage{latexsym}
\usepackage{graphicx}
\usepackage{booktabs}
\usepackage{CJK}
\usepackage{url}
\usepackage{amsfonts}
\usepackage{mathrsfs}
\usepackage{mathtools}
\usepackage{setspace}
\usepackage{amsmath}
\usepackage{multirow}
\usepackage{makecell,rotating}
\usepackage{array}
\aclfinalcopy 

\title{Duality Regularization for Unsupervised Bilingual Lexicon Induction}

\author{Xuefeng Bai$^{1,2,3}$,
  Yue Zhang$^{2,3}$,
  Hailong Cao$^4$,
  Tiejun Zhao$^4$ \\
  $^1$Zhejiang University \\
  $^2$School of Engineering, Westlake University\\
  $^3$Institute of Advanced Technology, Westlake Institute for Advanced Study \\
  $^4$Harbin Institute of Technology
  }
  
\begin{document}
\maketitle
\begin{abstract}
  Unsupervised bilingual lexicon induction  naturally exhibits duality, which results from symmetry in back-translation. For example, 
  EN-IT and IT-EN
  induction can be mutually primal and dual problems. 
  Current state-of-the-art methods, however, consider the two tasks independently.
  In this paper, we propose to 
  train primal and dual models jointly, using regularizers to encourage consistency in back translation cycles.
  Experiments across $6$ language pairs show that the proposed method significantly outperforms competitive baselines, obtaining the best published results on a standard benchmark.
\end{abstract}

\section{Introduction}

Unsupervised  bilingual lexicon induction (UBLI) 
has been shown to benefit NLP tasks for low resource languages, including unsupervised NMT~\cite{artetxe18EMNLP,artetxe18ICLR,Yang18ACL,lample18ICLR,Lample18EMNLP}, information retrieval~\cite{Vulic15,Litschko18}, dependency parsing~\cite{Guo15ACL}, and named entity recognition~\cite{mayhew17,xie18emnlp}.

Recent research has attempted to induce bilingual lexicons by aligning monolingual word vector spaces~\cite{Zhang17a,Conneau18a,aldarmaki18TACL,Artetxe18,Alvarez18EMNLP,Mukherjee18EMNLP,Bai18}. 
Given a pair of languages, their word alignment is inherently a bi-directional problem (e.g. English-Italian vs Italian-English). However, most existing research considers mapping from one language to another without making use of symmetry.
Our experiments show that separately learned UBLI models are not always consistent in opposite directions.
As shown in Figure~1a, when the model of \citeauthor{Conneau18a}~\shortcite{Conneau18a} is applied to English and Italian, the primal model maps the word ``three'' to the Italian word ``tre'', but the dual model maps ``tre'' to ``two'' instead of ``three''.

\begin{figure}
  \setlength{\abovecaptionskip}{0.2cm}   
  \setlength{\belowcaptionskip}{-0.6cm} 
  \centering\includegraphics[width=0.5\textwidth]{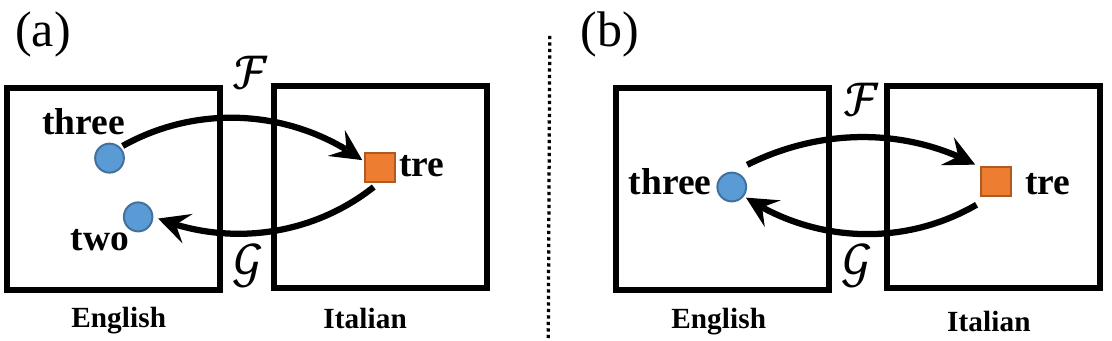}
  \footnotesize{\caption{(a) 
  Inconsistency between primal model $\mathcal{F}$ and the dual model $\mathcal{G}$. (b) An ideal scenario.}}  
  \label{fig:zero}
\end{figure}

We propose to address this issue by exploiting duality, encouraging forward and backward mappings to form a closed loop (Figure 1b).
In particular, we extend the model of ~\citeauthor{Conneau18a}~\shortcite{Conneau18a} by using a cycle consistency loss~\cite{Zhou16} to regularize two models in opposite directions. 
Experiments on two benchmark datasets show that the simple method of enforcing consistency gives better results in both directions. Our model significantly outperforms competitive baselines, obtaining the best-published results. 
We release our code at xxx.

\section{Related Work}
\textbf{UBLI}.
A typical line of work uses adversarial training~\cite{barone16,Zhang17a,Zhang17b,Conneau18a,bai-etal-19-BiAAE},
matching the distributions of source and target word embeddings through generative adversarial networks~\cite{Goodfellow14}.
Non-adversarial approaches have also been explored. 
For instance, ~\citeauthor{Mukherjee18EMNLP}~\shortcite{Mukherjee18EMNLP} use squared-loss mutual information to search for optimal cross-lingual word pairing. 
~\citet{Artetxe18} and ~\citet{Hoshen18} exploit the structural similarity of word embedding spaces to learn word mappings.
In this paper, we choose \citeauthor{Conneau18a}~\shortcite{Conneau18a} as our baseline as it is theoretically attractive and gives strong results on large-scale datasets. 

\noindent\textbf{Cycle Consistency.}
Forward-backward consistency has been used to discover the correspondence between unpaired images~\cite{Zhu17,Kim17}.
In machine translation, similar ideas were exploited, ~\citet{He16},
~\citet{Xia17ijcai}
and \citet{Wang18aaai} use dual learning to train two “opposite” language translators by minimizing the reconstruction loss. ~\citet{Sennrich16} consider back-translation, where a backward model is used to build  synthetic parallel corpus and a forward model learns to generate genuine text based on the synthetic output. 

Closer to our method, ~\citet{Chandar14} jointly train two autoencoders to learn supervised bilingual word embeddings. ~\citet{Xu18EMNLP} 
use sinkhorn distance~\cite{Marco13} and back-translation to align word embeddings.
However, they cannot perform fully unsupervised training, relying on WGAN~\cite{arjovsky17} for providing initial mappings. 
Concurrent with our work,  ~\citet{mohiuddin19} build a adversarial autoencoder with cycle consistency loss and post-cycle reconstruction loss. 
In contrast to these works, our method is fully unsupervised, simpler, and empirically more effective.

\section{Approach}
We take \citet{Conneau18a} as our baseline, introducing a novel regularizer to enforce cycle consistency.
Let $X=\{x_1,...,x_n\}$ and $Y=\{y_1,...,y_m\}$ be two sets of $n$ and $m$ word embeddings for a source and a target language, respectively. The primal UBLI task aims to learn a linear mapping $\mathcal{F}:X\to Y$ such that for each $x_i$, $\mathcal{F}(x_i)$ corresponds to its
translation in $Y$. Similarly, a linear mapping $\mathcal{G}:Y\to X$ is defined for the dual task. In addition, we introduce two language discriminators $D_x$ and $D_y$, which are trained to discriminate between the mapped word embeddings and the original word embeddings.

\subsection{Baseline Adversarial Model}
\citet{Conneau18a} align two word embedding spaces through generative adversarial networks, 
in which two networks are trained simultaneously.
Specifically, take the primal UBLI task as an example, the linear mapping $\mathcal{F}$ tries to generate ``fake'' word embeddings $\mathcal{F}(x)$ that look similar to word embeddings from $Y$, while the discriminator $D_y$ aims to distinguish between ``fake'' and real word embeddings from $Y$. 
Formally, this idea can be expressed as the min-max game min$_{\mathcal{F}}$max$_{D{_y}}\ell_{adv}(\mathcal{F},D_y,X,Y)$, where
\begin{equation*}
 \ell_{adv}(\mathcal{F},D_y,X,Y) = \frac{1}{m}\sum\limits_{j=1}^{m}\textup{log} P_{D_y}(src=1|y_j)
\end{equation*}
\begin{equation}
+ \frac{1}{n}\sum\limits_{i=1}^n\textup{log}P_{D_y}(src=0|\mathcal{F}(x_i)).
\end{equation}
$P_{D_y}(src|y_j)$ is a model probability from $D_y$ to distinguish whether word embedding $y_j$ is coming from the target language (src = 1) or the primal mapping $\mathcal{F}$ (src = 0).
Similarly, the {dual} UBLI problem can be formulated as min$_{\mathcal{G}}$max$_{D_x}\ell_{adv}(\mathcal{G},D_x,Y,X)$, where $\mathcal{G}$ is the dual mapping, and $D_x$ is a source discriminator.

Theoretically, a unique solution for above minmax game exists, with the mapping and the discriminator reaching a nash equilibrium.
Since the adversarial training happens at the   distribution level, no cross-lingual supervision is required.

\begin{figure}[t]
  \centering\includegraphics[width=0.5\textwidth]{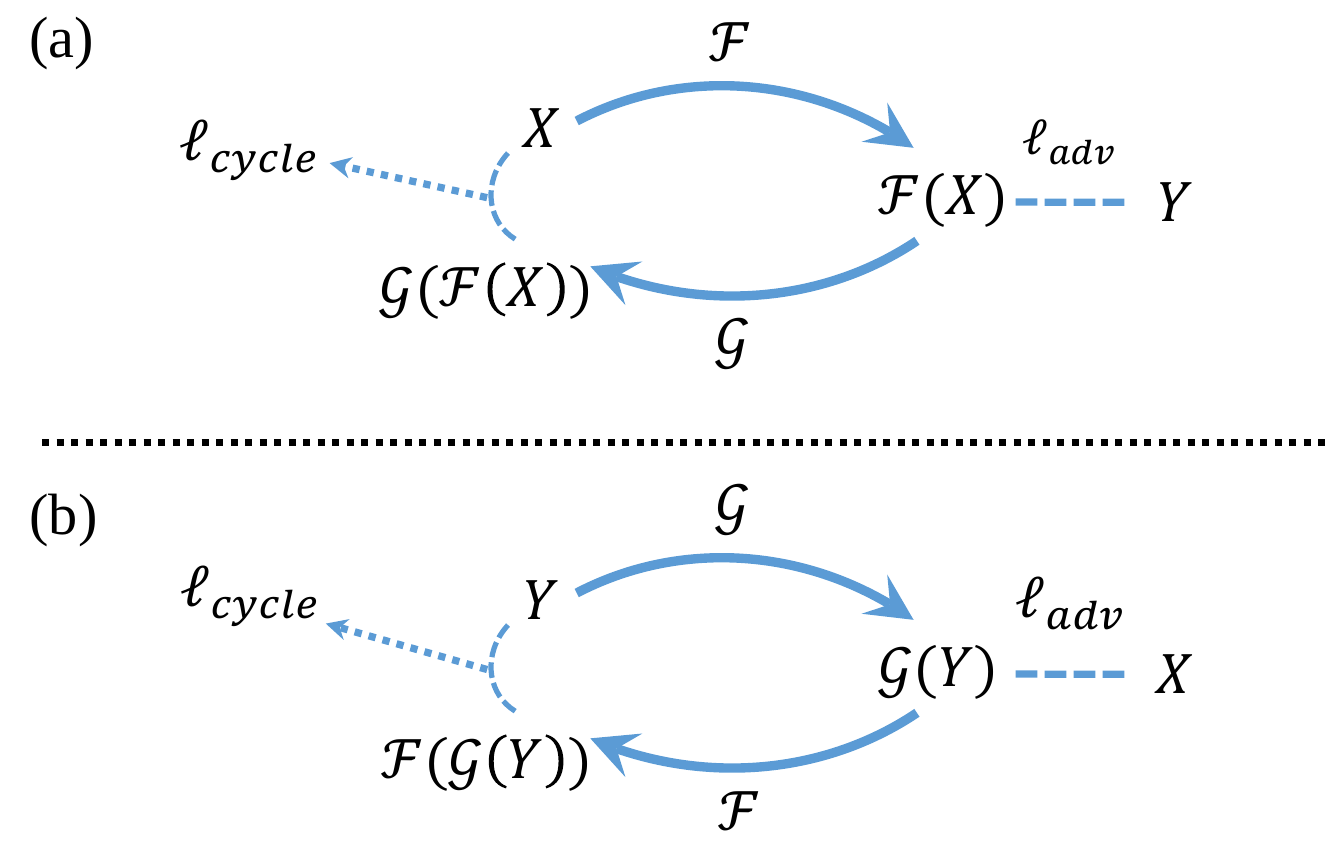}
 \footnotesize\caption{The proposed framework. 
 (a) $X\to\mathcal{F}(X)\to\mathcal{G}(\mathcal{F}(X))\to X$; (b) $Y\to\mathcal{G}(Y)\to\mathcal{F}(\mathcal{G}(Y))\to Y$.}  
  \label{fig:one}
\end{figure}
\subsection{Regularizers for Dual Models}
We train $\mathcal{F}$ and $\mathcal{G}$ jointly and introduce two regularizers.
Formally, we hope that $\mathcal{G}(\mathcal{F}(X))$ is similar to $X$ and $\mathcal{F}(\mathcal{G}(Y))$ is similar to $Y$.
We implement this constraint as a cycle consistency loss.
As a result, the proposed model has two learning objectives: i) an adversarial loss ($\ell_{adv}$) for each model as in the baseline.
ii) a cycle consistency loss ($\ell_{cycle}$) on each side to avoid $\mathcal{F}$ and $\mathcal{G}$ from contradicting each other. The overall architecture  of  our  model  is  illustrated  in Figure~\ref{fig:one}.



\noindent\textbf{Cycle Consistency Loss.} 
We introduce
\begin{equation}
\label{eqn:3}
\begin{split}
 &\ell_{cycle}(\mathcal{F},\mathcal{G},X) = 
 \frac{1}{n}\sum\limits_{i=1}^{n}\Delta(x_i ,\mathcal{G}(\mathcal{F}(x_i))),\\
 &\ell_{cycle}(\mathcal{F},\mathcal{G},Y) = 
 \frac{1}{m}\sum\limits_{j=1}^{m}\Delta(y_j,\mathcal{F}(\mathcal{G}(y_j))),\\
\end{split}
\end{equation}
 
\noindent where $\Delta$ denotes the discrepancy criterion, which is set as the average cosine similarity in our model.

\noindent\textbf{Full objective.} The final objective is:
\begin{equation*}
\begin{split}
\ell(\mathcal{F},&\mathcal{G},D_x,D_y,X,Y) = 
 \\&\ell_{adv}(\mathcal{F},D_y,X,Y)+\ell_{adv}(\mathcal{G},D_x,Y,X)
 \end{split}
\end{equation*}
\begin{equation}
\label{eqn:5}
 +\ell_{cycle}(\mathcal{F},\mathcal{G},X)+\ell_{cycle}(\mathcal{F},\mathcal{G},Y).
\end{equation}
\subsection{Model Selection}
We follow ~\citet{Conneau18a}, using an unsupervised criterion to perform model selection.
In preliminary experiments, we find in adversarial training that the single-direction criterion $S(\mathcal{F}, X, Y)$ by~\citet{Conneau18a} does not always work well.
To address this, we make a simple extension by calculating the weighted average of forward and backward scores:
\begin{equation}
\label{eqn:6}
\begin{split}
 S_{a}= \lambda S(\mathcal{F}, X, Y) + (1-\lambda)S(\mathcal{G}, X, Y),
\end{split}
\end{equation}
Where $\lambda$ is a hyperparameter to control the importance of the two objectives.\footnote{We find that $\lambda=0.5$ generally works well.}
Here $S$ first generates bilingual lexicons by learned mappings, and then computes the average cosine similarity of these translations.

\section{Experiments}
We perform two sets of experiments, to investigate the effectiveness of our duality regularization in isolation (Section~\ref{sec42}) and to compare our final models with the state-of-the-art methods in the literature (Section~\ref{sec43}), respectively.

\subsection{Experimental Settings}

\noindent\textbf{Dataset and Setup}. 
Our datasets includes: (i) The Multilingual Unsupervised and Supervised Embeddings {(\textbf{MUSE})} dataset released by \citeauthor{Conneau18a}~\shortcite{Conneau18a}. 
(ii) the more challenging \textbf{Vecmap} dataset from~\citeauthor{Dinu15}~\shortcite{Dinu15} and the extensions of \citeauthor{Artetxe17ACL}~\shortcite{Artetxe17ACL}. We follow the evaluation setups of ~\citet{Conneau18a}, utilizing cross-domain similarity local scaling (CSLS) for retrieving the translation of given source words. Following a standard evaluation practice ~\cite{vulic13,Mikolov13b,Conneau18a}, we report precision at 1 scores (P@1).
Given the instability of existing methods, we follow \citet{Artetxe18} to perform 10 runs for each method and report the best and the average accuracies.
\begin{table}[t]
\begin{centering}
\scalebox{0.85}{
\begin{tabular}{|l|c|cccc|}
\hline 
 \multicolumn{2}{|c|}{\multirow{2}{*}{Setting}}  & \multicolumn{2}{c} {Adv-C} &\multicolumn{2}{c|} {Ours} \tabularnewline 
\cline{3-6}
 \multicolumn{2}{|c|}{} & best & average. &best &average.  \tabularnewline
\hline
\multirow{10}{*}{\begin{sideways}\textbf{MUSE}\end{sideways}}& EN-ES  &77.3 &75.1 &\textbf{78.4} &\textbf{77.0}\tabularnewline
& ES-EN  &\textbf{79.1} &73.5 &79.0 &\textbf{75.6} \tabularnewline
\cline{2-6}
& EN-DE  &69.2 &32.4 &\textbf{70.0} &\textbf{56.5} \tabularnewline
& DE-EN &68.5 &31.7 &\textbf{69.3} &\textbf{53.7} \tabularnewline
\cline{2-6}
& EN-IT  &65.2 &47.7 &\textbf{72.0} &\textbf{71.1} \tabularnewline
& IT-EN  &64.0 &45.3 &\textbf{69.9} &\textbf{69.4} \tabularnewline
\cline{2-6}
& EN-EO  &18.6 &13.5 &\textbf{20.9} &\textbf{17.5}\tabularnewline
& EO-EN  &16.6 &12.0 &\textbf{17.3} &\textbf{15.3} \tabularnewline
\cline{2-6}
& EN-MS  &17.9 &08.3 &\textbf{24.7} &\textbf{21.8}\tabularnewline
& MS-EN &19.2 &06.4 &\textbf{27.6} &\textbf{23.5} \tabularnewline
\hline 
\multirow{8}{*}{\begin{sideways}\textbf{Vecmap}\end{sideways}}& EN-ES  &26.2 &20.5 &\textbf{29.6} &\textbf{26.1}\tabularnewline
& ES-EN &00.0 &00.0 &\textbf{21.7} &\textbf{20.2}\tabularnewline
\cline{2-6}
& EN-DE &40.3 &20.0 &\textbf{43.7} &\textbf{36.5}\tabularnewline
& DE-EN &00.0 &00.0 &\textbf{37.8} &\textbf{33.4}\tabularnewline
\cline{2-6}
& EN-IT &38.3 &37.0 & \textbf{38.5} &\textbf{37.5} \tabularnewline
& IT-EN &33.6 &14.7 &\textbf{34.7} &\textbf{33.1}\tabularnewline
\cline{2-6}
& EN-FI &01.9 &00.3 &\textbf{22.2} &\textbf{21.9}\tabularnewline
& FI-EN &00.0 &00.0 &\textbf{20.0} &\textbf{18.9}\tabularnewline
\hline 
\end{tabular}}
\par\end{centering}
\footnotesize\caption{\label{tab:01} Accuracy on MUSE and Vecmap.}
\end{table}

\begin{table}[htbp]
\begin{centering}
\scalebox{0.8}{
\begin{tabular}{lccccc}
\hline 
 &EN-ES & EN-DE & EN-IT & EN-EO &EN-MS  \tabularnewline
\hline 
Adv-C &66.95\% &67.83\% &70.23\% &72.30\% &75.87\% \tabularnewline
Ours&63.58\% &64.29\% &65.05\% &64.06\% &68.84\% \tabularnewline
\hline
\end{tabular}}
\par\end{centering}
\footnotesize\caption{\label{tab:02} Inconsistency rates on MUSE.}
\end{table}
\begin{table}[!ht]
\vspace{-0cm}
\begin{centering}
\scalebox{0.9}{
\begin{tabular}{cc}
\hline 
\multicolumn{1}{c} {Adv-C} &\multicolumn{1}{c} {Ours} \tabularnewline 
\hline 
\textbf{three-tre}-two &\textbf{three-tre-three}\tabularnewline
\textbf{neck-collo}-ribcage
&\textbf{neck-collo-neck}\tabularnewline
door-\textbf{finestrino-window}&\textbf{door-portiera-door}\tabularnewline
second-\textbf{terzo-third} &second-terzo-second\tabularnewline
before-\textbf{prima-first} & before-\textbf{dopo-after}\tabularnewline
\hline
\end{tabular}}
\par\end{centering}
\footnotesize\caption{\label{tab:03}Word translation examples for English-Italian on MUSE. Ground truths are marked in \textbf{bold}.}
\end{table}
\begin{table*}[t]
\begin{centering}
\scalebox{0.85}{\begin{tabular}{clcccccccc}
\hline 
\multirow{2}{*}{Supervision} &\multirow{2}{*}{Approach}& \multicolumn{2}{c} {EN-IT} & \multicolumn{2}{c} {EN-DE}& \multicolumn{2}{c} {EN-FI}& \multicolumn{2}{c} {EN-ES}\tabularnewline 
 & & $\rightarrow$ & $\leftarrow$ &$\rightarrow$ & $\leftarrow$ &$\rightarrow$ & $\leftarrow$ & $\rightarrow$ & $\leftarrow$  \tabularnewline
\hline
\multirow{4}{*}{\shortstack{Supervised\\ Methods}} & Procrustes & 45.33 & 39.05 & 47.27 &41.13  &32.16 &30.01 &36.67 &30.94 \tabularnewline
& GPA$\dagger$ & 45.33 & - & 48.46 &- &31.39 &- &- &- \tabularnewline
& GeoMM & 48.17 & 41.10 & 49.40 & 44.73 &36.03 &38.24 &39.27 &34.58 \tabularnewline
& GeoMM$_{semi}$ & \textbf{50.00} & \textbf{42.67} & 51.47 & 46.96 &\textbf{36.24} &39.57 &39.30 &36.06 \tabularnewline
\hline
\multirow{5}{*}{\shortstack{Unsupervised\\ Methods}}& Adv-C-Procrustes & 45.40 &38.78 & 46.40 & 00.00 &25.21 &00.15 &35.47 &0.05 \tabularnewline
& Unsup-SL & 48.01 & 42.10 & 48.22 &44.09  &32.95 &33.45 &37.47 &31.59 \tabularnewline
& Sinkhorn-BT & 44.67 & 38.77 & 44.53 & 41.93  & 23.53 & 23.42 & 32.13 & 27.62 \tabularnewline
& Ours-Procrustes &45.60  &38.29  &46.58  &42.50  &28.08  &26.48  &35.20  &28.94  \tabularnewline
& Ours-GeoMM$_{semi}$ &\textbf{50.00} &\textbf{42.67} &\textbf{51.60} &\textbf{47.22} &35.88 &\textbf{39.62}  &\textbf{39.47} &\textbf{36.43}
\tabularnewline
\hline \end{tabular}}
\par\end{centering}
\caption{\label{tab:04} Accuracy (P@1) on \textbf{Vecmap}. The best results are \textbf{bolded}. $\dagger$Results as reported in the original paper. For unsupervised methods, we report the average accuracy across 10 runs.}
\end{table*}
\subsection{The Effectiveness of Dual Learning}
\label{sec42}
We compare our method with ~\citet{Conneau18a} (Adv-C)  under the same settings. 
As shown in Table~\ref{tab:01}, our model outperforms Adv-C on both MUSE and Vecmap for all language pairs (except ES-EN).
In addition, the proposed approach is less sensitive to initialization, and thus more stable than Adv-C over multiple runs.
These results demonstrate the effectiveness of dual learning.
Our method is also superior to Adv-C for the low-resource language pairs English $\leftrightarrow$ Malay (MS) and English $\leftrightarrow$ English-Esperanto (EO).
Adv-C gives low performances on ES-EN, DE-EN, but much better results on the opposite directions on Vecmap. 
This is likely because the separate models are highly under-constrained, and thus easy to get stuck in poor local optima.
In contrast, our method gives comparable results on both directions for the two languages, thanks to the use of information symmetry.

Table~\ref{tab:02} shows the inconsistency rates\footnote{For each word $x_i$ from the source language, we check whether the primal $\mathcal{F}$ and the dual mapping $\mathcal{G}$ can recover $x_i$, i.e. $x_i\rightarrow\mathcal{F}(x_i) \rightarrow\mathcal{G}(\mathcal{F}(x_i))\rightarrow x_i$.} of back translation between Adv-C and our method on MUSE. Compared with Adv-C, our model significantly reduces the inconsistency rates on all language pairs, which explains the overall improvement in Table~\ref{tab:01}.
Table~\ref{tab:03} gives several word translation examples. In the first three cases, our regularizer successfully fixes back translation errors. In the fourth case, ensuring cycle consistency does not lead to the correct translation, which explains some errors by our system. In the fifth case, our model finds a related word but not the same word in the back translation, due to the use of cosine similarity for regularization.
\subsection{Comparison with the State-of-the-art}
\label{sec43}
In this section, we compare our model with state-of-the-art systems, including
those with different degrees of supervision.
The baselines include: (1) \textbf{Procrustes}~\cite{Conneau18a}, which learns a linear mapping through Procrustes Analysis~\cite{Schonemann1966}. (2)~\textbf{GPA}~\cite{Kementchedjhieva18}, an extension of  Procrustes Analysis. (3)~\textbf{ GeoMM}~\cite{Jawanpuria19}, a geometric approach which learn a Mahalanobis metric to refine the notion of similarity. (4)~\textbf{GeoMM$_{semi}$}, iterative GeoMM with weak supervision. (5) \textbf{Adv-C-Procrustes}~\cite{Conneau18a}, which refines the mapping learned by Adv-C with iterative Procrustes, which learns the new mapping matrix by constructing a bilingual lexicon iteratively.
(6) \textbf{Unsup-SL}~\cite{Artetxe18}, which integrates a weak unsupervised mapping with a robust self-learning.
(7) \textbf{Sinkhorn-BT}~\cite{Xu18EMNLP}, which combines sinkhorn distance~\cite{Marco13} and  back-translation.
For fair comparison, we integrate our model with two iterative refinement methods (Procrustes and GeoMM$_{semi}$).

Table~\ref{tab:04} shows the final results on Vecmap.\footnote{We select Vecmap as it is more challenging and closer to the real scenarios than MUSE~\cite{Artetxe18}.}
We first compare our model with the state-of-the-art unsupervised methods.
Our model based on procrustes (Ours-Procrustes) outperforms Sinkhorn-BT on all test language pairs, and shows better performance than Adv-C-Procrustes on most language pairs.  
Adv-C-Procrustes gives very low precision on DE-EN, FI-EN and ES-EN, while Ours-Procrustes obtains reasonable results consistently. 
A possible explanation is that dual learning is helpful for providing good initiations, so that the procrustes solution is not likely to fall in  poor local optima.
The reason why Unsup-SL gives strong results on all language pairs is that it uses a robust self-learning framework, which contains several techniques to avoid poor local optima.

Additionally, we observe that our unsupervised method performs competitively and even better compared with strong supervised and semi-supervised approaches. 
Ours-Procrustes obtains comparable results with Procrustes on EN-IT and gives strong results on EN-DE, EN-FI, EN-ES and the opposite directions.
Ours-GeoMM$_{semi}$ obtains the state-of-the-art results on all tested language pairs except EN-FI, with the additional advantage of being fully unsupervised. 
\section{Conclusion}
We investigated a regularization method to enhance unsupervised bilingual lexicon induction, by encouraging symmetry in lexical mapping between a pair of word embedding spaces. Results show that strengthening bi-directional mapping consistency significantly improves the effectiveness over the state-of-the-art method, leading to the best results on a standard benchmark.





\bibliographystyle{acl_natbib}
\bibliography{custom}

\begin{thebibliography}{41}
\expandafter\ifx\csname natexlab\endcsname\relax\def\natexlab#1{#1}\fi

\bibitem[{Aldarmaki et~al.(2018)Aldarmaki, Mohan, and Diab}]{aldarmaki18TACL}
Hanan Aldarmaki, Mahesh Mohan, and Mona Diab. 2018.
\newblock \href {http://aclweb.org/anthology/Q18-1014} {Unsupervised word
  mapping using structural similarities in monolingual embeddings}.
\newblock \emph{Transactions of the Association for Computational Linguistics},
  6:185--196.

\bibitem[{Alvarez-Melis and Jaakkola(2018)}]{Alvarez18EMNLP}
David Alvarez-Melis and Tommi Jaakkola. 2018.
\newblock \href {http://aclweb.org/anthology/D18-1214} {Gromov-wasserstein
  alignment of word embedding spaces}.
\newblock In \emph{Proceedings of the 2018 Conference on Empirical Methods in
  Natural Language Processing}, pages 1881--1890. Association for Computational
  Linguistics.

\bibitem[{Arjovsky et~al.(2017)Arjovsky, Chintala, and Bottou}]{arjovsky17}
Mart{\'i}n Arjovsky, Soumith Chintala, and L{\'e}on Bottou. 2017.
\newblock Wasserstein gan.
\newblock \emph{arXiv preprint arXiv:1701.07875}.

\bibitem[{Artetxe et~al.(2017)Artetxe, Labaka, and Agirre}]{Artetxe17ACL}
Mikel Artetxe, Gorka Labaka, and Eneko Agirre. 2017.
\newblock \href {https://doi.org/10.18653/v1/P17-1042} {Learning bilingual word
  embeddings with (almost) no bilingual data}.
\newblock In \emph{Proceedings of the 55th Annual Meeting of the Association
  for Computational Linguistics (Volume 1: Long Papers)}, pages 451--462.
  Association for Computational Linguistics.

\bibitem[{Artetxe et~al.(2018{\natexlab{a}})Artetxe, Labaka, and
  Agirre}]{Artetxe18}
Mikel Artetxe, Gorka Labaka, and Eneko Agirre. 2018{\natexlab{a}}.
\newblock \href {http://aclweb.org/anthology/P18-1073} {A robust self-learning
  method for fully unsupervised cross-lingual mappings of word embeddings}.
\newblock In \emph{Proceedings of the 56th Annual Meeting of the Association
  for Computational Linguistics (Volume 1: Long Papers)}, pages 789--798.
  Association for Computational Linguistics.

\bibitem[{Artetxe et~al.(2018{\natexlab{b}})Artetxe, Labaka, and
  Agirre}]{artetxe18EMNLP}
Mikel Artetxe, Gorka Labaka, and Eneko Agirre. 2018{\natexlab{b}}.
\newblock \href {http://aclweb.org/anthology/D18-1399} {Unsupervised
  statistical machine translation}.
\newblock In \emph{Proceedings of the 2018 Conference on Empirical Methods in
  Natural Language Processing}, pages 3632--3642. Association for Computational
  Linguistics.

\bibitem[{Artetxe et~al.(2018{\natexlab{c}})Artetxe, Labaka, Agirre, and
  Cho}]{artetxe18ICLR}
Mikel Artetxe, Gorka Labaka, Eneko Agirre, and Kyunghyun Cho.
  2018{\natexlab{c}}.
\newblock \href {https://openreview.net/forum?id=Sy2ogebAW} {Unsupervised
  neural machine translation}.
\newblock In \emph{International Conference on Learning Representations}.

\bibitem[{Bai et~al.(2019)Bai, Cao, Chen, and Zhao}]{bai-etal-19-BiAAE}
Xuefeng Bai, Hailong Cao, Kehai Chen, and Tiejun Zhao. 2019.
\newblock \href {https://doi.org/10.1109/TASLP.2019.2925973} {A bilingual
  adversarial autoencoder for unsupervised bilingual lexicon induction}.
\newblock \emph{IEEE/ACM Transactions on Audio, Speech, and Language
  Processing}, 27(10):1639--1648.

\bibitem[{Bai et~al.(2018)Bai, Cao, and Zhao}]{Bai18}
Xuefeng Bai, Hailong Cao, and Tiejun Zhao. 2018.
\newblock \href {http://doi.acm.org/10.1145/3197566} {Improving vector space
  word representations via kernel canonical correlation analysis}.
\newblock \emph{ACM Transactions on Asian and Low-Resource Language Information
  Processing}, 17(4):29:1--29:16.

\bibitem[{Chandar et~al.(2014)Chandar, Lauly, Larochelle, Khapra, Ravindran,
  Raykar, and Saha}]{Chandar14}
A.~P.~Sarath Chandar, Stanislas Lauly, Hugo Larochelle, Mitesh~M. Khapra,
  Balaraman Ravindran, Vikas~C. Raykar, and Amrita Saha. 2014.
\newblock An autoencoder approach to learning bilingual word representations.
\newblock In \emph{NIPS}.

\bibitem[{Conneau et~al.(2018)Conneau, Lample, Ranzato, Denoyer, and
  J{\'e}gou}]{Conneau18a}
Alexis Conneau, Guillaume Lample, Marc'Aurelio Ranzato, Ludovic Denoyer, and
  Herv{\'e} J{\'e}gou. 2018.
\newblock \href {https://openreview.net/forum?id=H196sainb} {Word translation
  without parallel data}.
\newblock In \emph{International Conference on Learning Representations}.

\bibitem[{Cuturi(2013)}]{Marco13}
Marco Cuturi. 2013.
\newblock \href
  {http://papers.nips.cc/paper/4927-sinkhorn-distances-lightspeed-computation-of-optimal-transport.pdf}
  {Sinkhorn distances: Lightspeed computation of optimal transport}.
\newblock In C.~J.~C. Burges, L.~Bottou, M.~Welling, Z.~Ghahramani, and K.~Q.
  Weinberger, editors, \emph{Advances in Neural Information Processing Systems
  26}, pages 2292--2300. Curran Associates, Inc.

\bibitem[{Dinu et~al.(2015)Dinu, Lazaridou, and Baroni}]{Dinu15}
Georgiana Dinu, Angeliki Lazaridou, and Marco Baroni. 2015.
\newblock Improving zero-shot learning by mitigating the hubness problem.
\newblock In \emph{Proceedings of the 3th International Conference on Learning
  Representations (ICLR)}.

\bibitem[{Goodfellow et~al.(2014)Goodfellow, Pouget-Abadie, Mirza, Xu,
  Warde-Farley, Ozair, Courville, and Bengio}]{Goodfellow14}
Ian~J. Goodfellow, Jean Pouget-Abadie, Mehdi Mirza, Bing Xu, David
  Warde-Farley, Sherjil Ozair, Aaron~C. Courville, and Yoshua Bengio. 2014.
\newblock Generative adversarial nets.
\newblock In \emph{NIPS}.

\bibitem[{Guo et~al.(2015)Guo, Che, Yarowsky, Wang, and Liu}]{Guo15ACL}
Jiang Guo, Wanxiang Che, David Yarowsky, Haifeng Wang, and Ting Liu. 2015.
\newblock \href {http://www.aclweb.org/anthology/P15-1119} {Cross-lingual
  dependency parsing based on distributed representations}.
\newblock In \emph{Proceedings of the 53rd Annual Meeting of the Association
  for Computational Linguistics and the 7th International Joint Conference on
  Natural Language Processing (Volume 1: Long Papers)}, pages 1234--1244,
  Beijing, China. Association for Computational Linguistics.

\bibitem[{He et~al.(2016)He, Xia, Qin, Wang, Yu, Liu, and Ma}]{He16}
Di~He, Yingce Xia, Tao Qin, Liwei Wang, Nenghai Yu, Tie-Yan Liu, and Wei-Ying
  Ma. 2016.
\newblock \href
  {http://papers.nips.cc/paper/6469-dual-learning-for-machine-translation.pdf}
  {Dual learning for machine translation}.
\newblock In D.~D. Lee, M.~Sugiyama, U.~V. Luxburg, I.~Guyon, and R.~Garnett,
  editors, \emph{Advances in Neural Information Processing Systems 29}, pages
  820--828. Curran Associates, Inc.

\bibitem[{Hoshen and Wolf(2018)}]{Hoshen18}
Yedid Hoshen and Lior Wolf. 2018.
\newblock \href {http://aclweb.org/anthology/D18-1043} {Non-adversarial
  unsupervised word translation}.
\newblock In \emph{Proceedings of the 2018 Conference on Empirical Methods in
  Natural Language Processing}, pages 469--478. Association for Computational
  Linguistics.

\bibitem[{Jawanpuria et~al.(2018)Jawanpuria, Balgovind, Kunchukuttan, and
  Mishra}]{Jawanpuria19}
Pratik Jawanpuria, Arjun Balgovind, Anoop Kunchukuttan, and Bamdev Mishra.
  2018.
\newblock \href {http://arxiv.org/abs/1808.08773} {Learning multilingual word
  embeddings in latent metric space: {A} geometric approach}.
\newblock \emph{CoRR}, abs/1808.08773.

\bibitem[{Kementchedjhieva et~al.(2018)Kementchedjhieva, Ruder, Cotterell, and
  S{\o}gaard}]{Kementchedjhieva18}
Yova Kementchedjhieva, Sebastian Ruder, Ryan Cotterell, and Anders S{\o}gaard.
  2018.
\newblock \href {http://aclweb.org/anthology/K18-1021} {Generalizing procrustes
  analysis for better bilingual dictionary induction}.
\newblock In \emph{Proceedings of the 22nd Conference on Computational Natural
  Language Learning}, pages 211--220. Association for Computational
  Linguistics.

\bibitem[{Kim et~al.(2017)Kim, Cha, Kim, Lee, and Kim}]{Kim17}
Taeksoo Kim, Moonsu Cha, Hyunsoo Kim, Jung~Kwon Lee, and Jiwon Kim. 2017.
\newblock Learning to discover cross-domain relations with generative
  adversarial networks.
\newblock In \emph{ICML}.

\bibitem[{Lample et~al.(2018{\natexlab{a}})Lample, Conneau, Denoyer, and
  Ranzato}]{lample18ICLR}
Guillaume Lample, Alexis Conneau, Ludovic Denoyer, and Marc'Aurelio Ranzato.
  2018{\natexlab{a}}.
\newblock \href {https://openreview.net/forum?id=rkYTTf-AZ} {Unsupervised
  machine translation using monolingual corpora only}.
\newblock In \emph{International Conference on Learning Representations}.

\bibitem[{Lample et~al.(2018{\natexlab{b}})Lample, Ott, Conneau, Denoyer, and
  Ranzato}]{Lample18EMNLP}
Guillaume Lample, Myle Ott, Alexis Conneau, Ludovic Denoyer, and Marc'Aurelio
  Ranzato. 2018{\natexlab{b}}.
\newblock \href {http://aclweb.org/anthology/D18-1549} {Phrase-based {\&}
  neural unsupervised machine translation}.
\newblock In \emph{Proceedings of the 2018 Conference on Empirical Methods in
  Natural Language Processing}, pages 5039--5049. Association for Computational
  Linguistics.

\bibitem[{Litschko et~al.(2018)Litschko, Glava\v{s}, Ponzetto, and
  Vuli\'{c}}]{Litschko18}
Robert Litschko, Goran Glava\v{s}, Simone~Paolo Ponzetto, and Ivan Vuli\'{c}.
  2018.
\newblock \href {https://doi.org/10.1145/3209978.3210157} {Unsupervised
  cross-lingual information retrieval using monolingual data only}.
\newblock In \emph{The 41st International ACM SIGIR Conference on Research
  \&\#38; Development in Information Retrieval}, SIGIR '18, pages 1253--1256,
  New York, NY, USA. ACM.

\bibitem[{Mayhew et~al.(2017)Mayhew, Tsai, and Roth}]{mayhew17}
Stephen Mayhew, Chen-Tse Tsai, and Dan Roth. 2017.
\newblock \href {https://doi.org/10.18653/v1/D17-1269} {Cheap translation for
  cross-lingual named entity recognition}.
\newblock In \emph{Proceedings of the 2017 Conference on Empirical Methods in
  Natural Language Processing}, pages 2536--2545. Association for Computational
  Linguistics.

\bibitem[{Miceli~Barone(2016)}]{barone16}
Antonio~Valerio Miceli~Barone. 2016.
\newblock \href {https://doi.org/10.18653/v1/W16-1614} {Towards cross-lingual
  distributed representations without parallel text trained with adversarial
  autoencoders}.
\newblock In \emph{Proceedings of the 1st Workshop on Representation Learning
  for NLP}, pages 121--126. Association for Computational Linguistics.

\bibitem[{Mikolov et~al.(2013)Mikolov, Le, and Sutskever}]{Mikolov13b}
Tomas Mikolov, Quoc~V. Le, and Ilya Sutskever. 2013.
\newblock \href {http://arxiv.org/abs/1309.4168} {Exploiting similarities among
  languages for machine translation}.
\newblock \emph{CoRR}, abs/1309.4168.

\bibitem[{Mohiuddin and Joty(2019)}]{mohiuddin19}
Tasnim Mohiuddin and Shafiq~R. Joty. 2019.
\newblock \href {http://arxiv.org/abs/1904.04116} {Revisiting adversarial
  autoencoder for unsupervised word translation with cycle consistency and
  improved training}.
\newblock \emph{CoRR}, abs/1904.04116.

\bibitem[{Mukherjee et~al.(2018)Mukherjee, Yamada, and
  Hospedales}]{Mukherjee18EMNLP}
Tanmoy Mukherjee, Makoto Yamada, and Timothy Hospedales. 2018.
\newblock \href {http://aclweb.org/anthology/D18-1063} {Learning unsupervised
  word translations without adversaries}.
\newblock In \emph{Proceedings of the 2018 Conference on Empirical Methods in
  Natural Language Processing}, pages 627--632. Association for Computational
  Linguistics.

\bibitem[{Sch{\"o}nemann(1966)}]{Schonemann1966}
Peter~H. Sch{\"o}nemann. 1966.
\newblock \href {https://doi.org/10.1007/BF02289451} {A generalized solution of
  the orthogonal procrustes problem}.
\newblock \emph{Psychometrika}, 31(1):1--10.

\bibitem[{Sennrich et~al.(2016)Sennrich, Haddow, and Birch}]{Sennrich16}
Rico Sennrich, Barry Haddow, and Alexandra Birch. 2016.
\newblock \href {https://doi.org/10.18653/v1/P16-1009} {Improving neural
  machine translation models with monolingual data}.
\newblock In \emph{Proceedings of the 54th Annual Meeting of the Association
  for Computational Linguistics (Volume 1: Long Papers)}, pages 86--96.
  Association for Computational Linguistics.

\bibitem[{Vuli{\'c} and Moens(2013)}]{vulic13}
Ivan Vuli{\'c} and Marie-Francine Moens. 2013.
\newblock Cross-lingual semantic similarity of words as the similarity of their
  semantic word responses.
\newblock In \emph{Proceedings of the 2013 Conference of the North American
  Chapter of the Association for Computational Linguistics: Human Language
  Technologies}, pages 106--116.

\bibitem[{Vuli\'{c} and Moens(2015)}]{Vulic15}
Ivan Vuli\'{c} and Marie-Francine Moens. 2015.
\newblock \href {https://doi.org/10.1145/2766462.2767752} {Monolingual and
  cross-lingual information retrieval models based on (bilingual) word
  embeddings}.
\newblock In \emph{Proceedings of the 38th International ACM SIGIR Conference
  on Research and Development in Information Retrieval}, SIGIR '15, pages
  363--372, New York, NY, USA. ACM.

\bibitem[{Wang et~al.(2018)Wang, Xia, Zhao, Bian, Qin, Liu, and
  Liu}]{Wang18aaai}
Yijun Wang, Yingce Xia, Lu~Zhao, Jiang Bian, Tao Qin, Guiquan Liu, and Tie-Yan
  Liu. 2018.
\newblock Dual transfer learning for neural machine translation with marginal
  distribution regularization.
\newblock In \emph{AAAI}.

\bibitem[{Xia et~al.(2017)Xia, Bian, Qin, Yu, and Liu}]{Xia17ijcai}
Yingce Xia, Jiang Bian, Tao Qin, Nenghai Yu, and Tie-Yan Liu. 2017.
\newblock \href {https://doi.org/10.24963/ijcai.2017/434} {Dual inference for
  machine learning}.
\newblock In \emph{Proceedings of the Twenty-Sixth International Joint
  Conference on Artificial Intelligence, {IJCAI-17}}, pages 3112--3118.

\bibitem[{Xie et~al.(2018)Xie, Yang, Neubig, Smith, and Carbonell}]{xie18emnlp}
Jiateng Xie, Zhilin Yang, Graham Neubig, Noah~A. Smith, and Jaime Carbonell.
  2018.
\newblock \href {https://arxiv.org/abs/1808.09861} {Neural cross-lingual named
  entity recognition with minimal resources}.
\newblock In \emph{Conference on Empirical Methods in Natural Language
  Processing (EMNLP)}, Brussels, Belgium.

\bibitem[{Xu et~al.(2018)Xu, Yang, Otani, and Wu}]{Xu18EMNLP}
Ruochen Xu, Yiming Yang, Naoki Otani, and Yuexin Wu. 2018.
\newblock \href {http://aclweb.org/anthology/D18-1268} {Unsupervised
  cross-lingual transfer of word embedding spaces}.
\newblock In \emph{Proceedings of the 2018 Conference on Empirical Methods in
  Natural Language Processing}, pages 2465--2474. Association for Computational
  Linguistics.

\bibitem[{Yang et~al.(2018)Yang, Chen, Wang, and Xu}]{Yang18ACL}
Zhen Yang, Wei Chen, Feng Wang, and Bo~Xu. 2018.
\newblock \href {http://aclweb.org/anthology/P18-1005} {Unsupervised neural
  machine translation with weight sharing}.
\newblock In \emph{Proceedings of the 56th Annual Meeting of the Association
  for Computational Linguistics (Volume 1: Long Papers)}, pages 46--55.
  Association for Computational Linguistics.

\bibitem[{Zhang et~al.(2017{\natexlab{a}})Zhang, Liu, Luan, and Sun}]{Zhang17a}
Meng Zhang, Yang Liu, Huanbo Luan, and Maosong Sun. 2017{\natexlab{a}}.
\newblock \href {https://doi.org/10.18653/v1/P17-1179} {Adversarial training
  for unsupervised bilingual lexicon induction}.
\newblock In \emph{Proceedings of the 55th Annual Meeting of the Association
  for Computational Linguistics (Volume 1: Long Papers)}, pages 1959--1970.
  Association for Computational Linguistics.

\bibitem[{Zhang et~al.(2017{\natexlab{b}})Zhang, Liu, Luan, and Sun}]{Zhang17b}
Meng Zhang, Yang Liu, Huanbo Luan, and Maosong Sun. 2017{\natexlab{b}}.
\newblock \href {https://doi.org/10.18653/v1/D17-1207} {Earth mover's distance
  minimization for unsupervised bilingual lexicon induction}.
\newblock In \emph{Proceedings of the 2017 Conference on Empirical Methods in
  Natural Language Processing}, pages 1934--1945. Association for Computational
  Linguistics.

\bibitem[{Zhou et~al.(2016)Zhou, Kr{\"a}henb{\"u}hl, Aubry, Huang, and
  Efros}]{Zhou16}
Tinghui Zhou, Philipp Kr{\"a}henb{\"u}hl, Mathieu Aubry, Qi-Xing Huang, and
  Alexei~A. Efros. 2016.
\newblock Learning dense correspondence via 3d-guided cycle consistency.
\newblock \emph{2016 IEEE Conference on Computer Vision and Pattern Recognition
  (CVPR)}, pages 117--126.

\bibitem[{Zhu et~al.(2017)Zhu, Park, Isola, and Efros}]{Zhu17}
Jun-Yan Zhu, Taesung Park, Phillip Isola, and Alexei~A. Efros. 2017.
\newblock Unpaired image-to-image translation using cycle-consistent
  adversarial networks.
\newblock In \emph{2017 IEEE International Conference on Computer Vision
  (ICCV)}, pages 2242--2251.

\end{thebibliography}
\end{document}